\documentclass[letterpaper, 10 pt, conference]{ieeeconf}  

\IEEEoverridecommandlockouts                              

\usepackage{cite}
\usepackage{hyperref}
\usepackage{amsmath,amssymb,amsfonts}
\usepackage{algorithmic}
\usepackage{graphicx}
\usepackage{color}
\usepackage{textcomp}
\usepackage{fourier} 
\usepackage{array}
\usepackage{makecell}
\usepackage{multirow}
\usepackage{balance}
\usepackage{tikz}

\newcommand\copyrighttext{%
  \footnotesize \textcopyright 2020 IEEE.  Personal use of this material is permitted.  Permission from IEEE must be obtained for all other uses, in any current or future media, including reprinting/republishing this material for advertising or promotional purposes, creating new collective works, for resale or redistribution to servers or lists, or reuse of any copyrighted component of this work in other works.}
\newcommand\copyrightnotice{%
\begin{tikzpicture}[remember picture,overlay]
\node[anchor=south,yshift=10pt] at (current page.south) {\fbox{\parbox{\dimexpr\textwidth-\fboxsep-\fboxrule\relax}{\copyrighttext}}};
\end{tikzpicture}%
}

\title{Keyframe-based Dense Mapping with the Graph\\ 
of View-Dependent Local Maps}

\author{Krzysztof Zieli\'nski$^{1}$, Dominik Belter$^{1}$
\thanks{*This work was supported by the National Centre for Research and Development (NCBR) through project LIDER/33/0176/L-8/16/NCBR/2017.}
\thanks{$^{1}$Institute of Control, Robotics and Information Engineering, Poznan University of Technology, Poznan, Poland
        {\tt\small dominik.belter@put.poznan.pl}}%
}

\begin{document}

\maketitle
\copyrightnotice
\thispagestyle{empty}
\pagestyle{empty}

\begin{abstract}
In this article, we propose a new keyframe-based mapping system. The proposed method updates local Normal Distribution Transform maps (NDT) using data from an RGB-D sensor. The cells of the NDT are stored in 2D view-dependent structures to better utilize the properties and uncertainty model of RGB-D cameras. This method naturally represents an object closer to the camera origin with higher precision. The local maps are stored in the pose graph which allows correcting global map after loop closure detection. We also propose a procedure that allows merging and filtering local maps to obtain a global map of the environment. Finally, we compare our method with Octomap and NDT-OM and provide example applications of the proposed mapping method.
\end{abstract}


\section{Introduction}
\noindent 
Autonomous robots use maps to model the environment. The model of the environment allows collision detection, motion planning, and localization. The map can also store information about the objects~\cite{Galvez2016} and geometrical primitives~\cite{Wietrzykowski2019} which can be found in the workspace of the robot. Feature-based SLAM systems like~\cite{Mur-Artal2017} store a sparse set of point features in the map. The point features are later used to re-localize the camera by matching features from the current camera image~\cite{Mur-Artal2017}. In contrast, dense mapping methods like Octomap~\cite{Hornung2013} or Normal Distribution Transform Occupancy Map (NDT-OM)~\cite{Saarinen2013} are focused on the global reconstruction of the objects and obstacles in the environment for collision detection and motion planning. In this research, we propose a new dense mapping method which is based on data structures used mainly by the sparse localization methods like ORB-SLAM2~\cite{Mur-Artal2017}.

\begin{figure}[t]
\centering
\includegraphics[width=0.69\columnwidth]{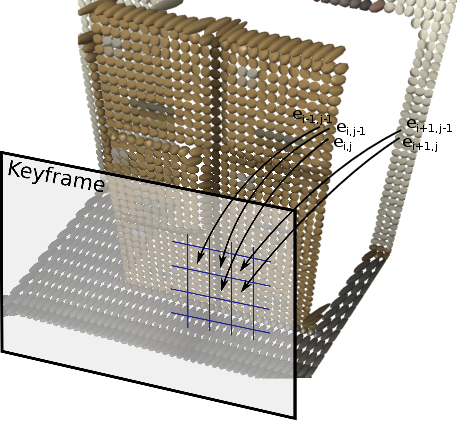}
\caption{Local map of the environment: normal distribution transforms (ellipsoids) projected on the image plane.}
\label{ideaLocalMap}
\end{figure}

The most popular methods to 3D dense mapping are Octomap~\cite{Hornung2013} and NDT-OM~\cite{Saarinen2013}. Both methods build a global map that divides the space into cells (voxels) and determines their occupancy. The size of the cells has to be determined in advance. It means that we can represent objects in the map with the given accuracy. The accuracy of the map depends on the tasks and it's different during grasping objects and planning the collision-free motion of the robot in the building. Also, decreasing the size of the cell increases significantly the number of cells. Moreover, the regular division of the space does not take into account the properties of the sensors (uncertainty model) during the update of the map and we lose precise information about the shape of the objects which are close to the robot. Finally, the sensor data is integrated into the global map taking into account current information about the pose of the robot. Whenever the robot re-localizes itself it corrects current and previous robot's poses~\cite{Mur-Artal2017}. However, with the global map, the dense measurements cannot be corrected because they are already integrated into the dense map.

In this paper, we propose a dense mapping method for static environments which is inspired by recent advances in robot localization and ORB-SLAM method. Instead of building a global map of the environment, we create a set of local maps (Fig.~\ref{ideaLocalMap}) which are related to the unique views from the camera (keyframes). We integrate depth measurements from the RGB-D sensor in the camera frame which allows representing objects closer to the camera with higher precision than distant objects. Despite the constant resolution in the local frame, we can represent the same regions of the 3D space with various precision and we don't lose information about details of the objects. Finally, we can generate a global map of the environment and correct the map after loop closure.

\begin{figure*}[t]
\centering
\includegraphics[width=0.89\textwidth]{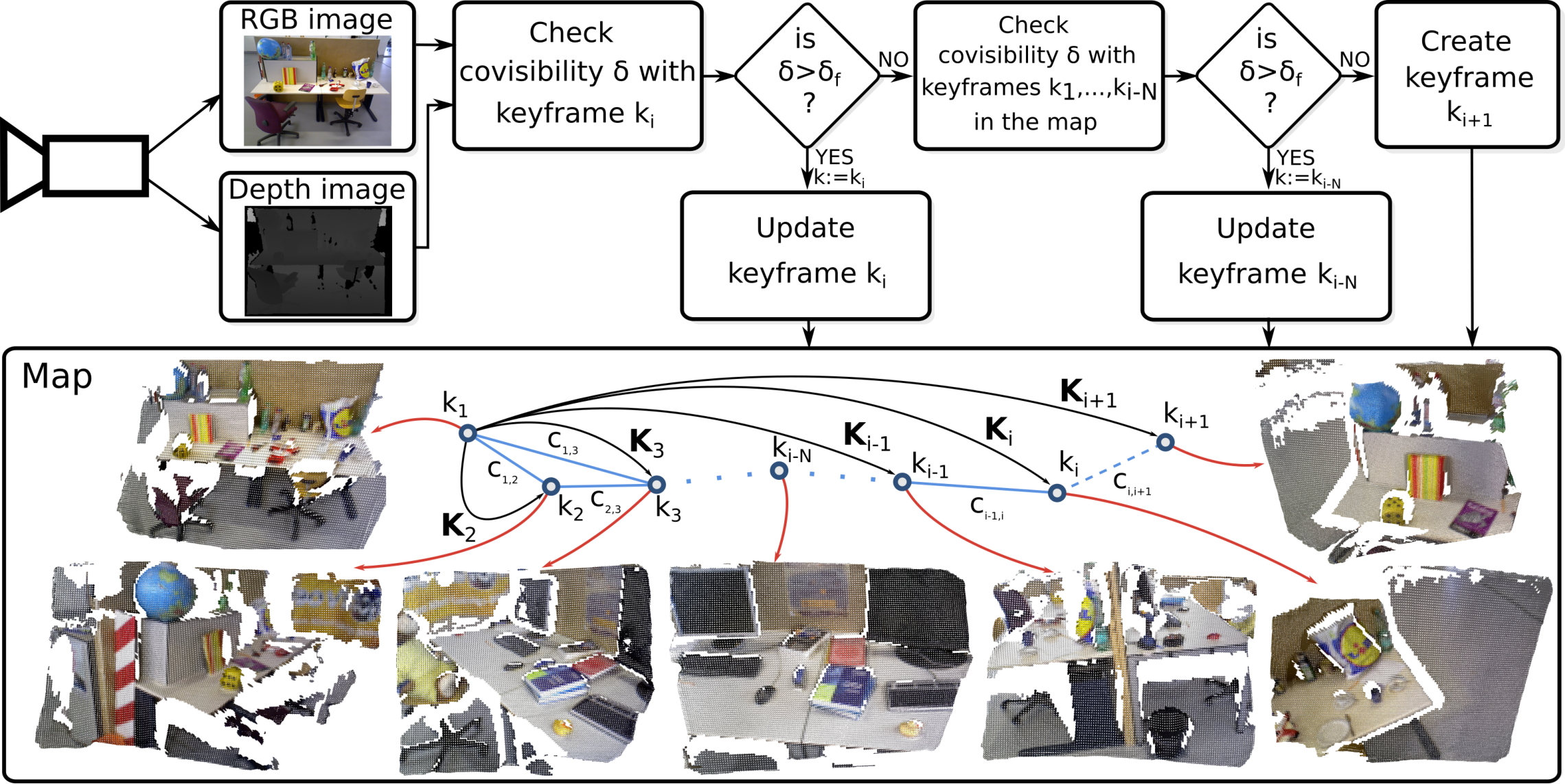}
\caption{Block diagram of the view-dependent keyframe-based mapping method}
\label{blockDiagram}
\end{figure*}

\subsection{Related Work}

The first implementations of the dense maps in robotics are 2.5D elevation maps~\cite{Krotkow1989,Kweon1992}. The terrain surface is obtained using the locus method~\cite{Kweon1992}. Moreover, data from multiple sensors are used to localize the robot using terrain matching. Recently, we presented that the 2.5D elevation map can be estimated using Kalman filtering to improve the accuracy of the map~\cite{Belter2016}. Fankhauser et al. suggest that the robot should be in the center of the local map which moves with the robot~\cite{Fankhauser2014}. This approach is justified by a growing uncertainty of the robot's position and decreasing the accuracy of objects which are far from the sensor. To tackle the problem of objects that are above the robot (such as tunnels) and avoid the complexity of 3D maps, extended multi-level elevation maps have been proposed by Pfaff et al.~\cite{Pfaff2007}.

One of the most popular approaches to 3D mapping of the environment in robotics is Octomap~\cite{Hornung2013}. Octomap stores occupied, free, and unknown spaces in the octree data structure. It is a memory-efficient method and guarantees logarithmic access time to each cell. Another algorithm that is based on Octree, updates corresponding parts of triangle-mesh~\cite{Steinbruecker2014}. This approach directly deals with the problem of representing the scene at multiple scales. Another approach to 3D mapping called multi-volume occupancy grid (MVOG) which explicitly stores information about the obstacles and free space reduces memory usage~\cite{Dryanovski2010}. Recently, dense 3D reconstruction and pose estimation~\cite{Dai2017} became computationally efficient. Also, multi-resolution maps consisting of surfels can be processed on-line using CPU only which is important for the most robotic applications~\cite{Stuckler}.

Plagemann et al.~\cite{Plagemann2008} suggested a new mapping approach as a regression problem using the Gaussian process. This terrain model has the advantage of predicting elevations at unseen locations more reliably than alternative approaches. Later, a similar approach was suggested by Saarinen et al.~\cite{Saarinen2013}. The proposed Normal Distribution Transform combines the advantages of both representations -- compactness of NDT-maps and robustness of occupancy maps. We also utilize this approach (NDT) to update 3D local maps.

Local maps (submap) are popular for the 2D models of the environment. This idea is implemented in the Google Cartographer SLAM~\cite{Hess2016}. The submaps are represented by probability grids connected by edges in the graph and later used for loop-closure. A similar graph-based approach was presented by Strom et al.~\cite{Strom2011} by rasterizing a dynamic occupancy grid. In 3D space, local maps and the concept of loop-closure was used in~\cite{Belter2018}. In contrast, ElasticFusion uses a dense surfel-based model obtained by dense frame-to-frame camera tracking without graph optimization~\cite{Wheelan2015}. Also, Ho et al. use Virtual Occupancy Grid Maps to correct global map after loop closure detection~\cite{Ho2018}. In this paper, we show the advantages of building 3D local maps and storing them in the pose-based graph. In~\cite{Meilland2013}, a similar approach is utilized but we use view-dependent representation as a local map which is memory efficient and better represents the uncertainty of the perception system. 

\subsection{Approach and Contribution}

In this paper, we use a 2D view-dependent approach to generate a dense model of the environment. As a result, the robot has a set of keyframes stored in the pose graph. Each keyframe contains information about the local model of the environment. The dense model is based on the NDT-OM but instead of dividing 3D space into voxels with the given size, we use the image plane to obtain 3D occupancy ellipsoids (normal distribution transforms). The discretization is performed on the image plane (Fig.~\ref{ideaLocalMap}).

The proposed structure of the map allows for correction of the global map after loop closure which is presented in our previous work~\cite{Belter2018}. In this research, we show that the proposed view-dependent model of the environment is more efficient than the 3D model (NDT-OM or Octomap) because it adapts the size of the cells of the map to the distance between the sensor (RGB-D camera) and the objects in the environment. Thus, the objects which are closer to the robot are represented in the map with higher precision than distant objects. Finally, the graph of local view-dependent dense maps allows restoring the global model of the environment.


\section{View-dependent Mapping}
\subsection{Global graph-based map}

The structure of the map and update procedure are presented in Fig.~\ref{blockDiagram}. The global map consists of a set of local maps (keyframes) $k_i$. The local maps are organized in a graph-like structure. The pose of each local map w.r.t. first keyframe is represented by the {\bf SE3} transformation ${\bf K}_i$. We also add edges that represent co-visibility between keyframes $\delta_{\rm i,j}$. The structure is inspired by the covisibility graph in the ORB-SLAM2~\cite{Mur-Artal2017}. The weight of the edge represents (covisibility $\delta$) the number of re-observed 2D features in the keyframes normalized by the total number of 2D features.

The update procedure of the map is presented in Fig.~\ref{blockDiagram}. We provide the pair of the RGB-D images to the input of the system to update the map. We assume that the camera pose is known and obtained from the independent localization system. First, we check the covisibility between the current RGB frame and the current keyframe $k_i$. If the covisiblility $\delta$ is above the threshold we update the current keyframe $k_i$. In other cases, we iterate over all keyframes in the map to detect loop closure. If the covisibility $\delta$ is above the threshold and we re-observe the scene which is stored in the map we update the keyframe which exists in the map $k_{\rm i-N}$. We create a new keyframe if the loop closure detection procedure does not recognize the current scene.

\subsection{Local maps - keyframes}
Opposed to other methods where 3D points are updated in 3D space, this method allows updating the data on a two-dimensional grid. First of all, we introduce a keyframe that consists of RGB-D pair of images, the 3D pose of the keyframe, a set of point features detected on the RGB image and a 2D container with ellipsoids (Fig.~\ref{ideaLocalMap}). Each ellipsoid is described by the 3D covariance matrix, 3D position ($[x,y,z]^T$), and color. The resolution of the container depends on the number of pixels used to update the ellipsoids in a single cell. In the experiments presented in the article, the size of each cell is set to 5$\times$5~px. Each 2D cell and position of the corresponding ellipsoid is updated using the RGB-D images and a color point cloud re-projected on the keyframe. The size of the 2D container is set to cover 1280$\times$960~px image. The RGB-D frame related to the keyframe is located in the center of the 2D container. Additionally, we show that the keyframe can contain information about objects detected on the RGB-D images.

\subsection{Update of the local map}

The color point cloud obtained from the RGB-D images and transformed to the pose of the considered keyframe is used to update the local map. Then, we use NDT update methodology~\cite{Saarinen2013} to obtain the position, color, and covariance matrix describing an ellipsoid. First, we compute a transformation matrix ${\bf T}$ between the keyframe and current camera pose

\begin{equation}
    {\bf T}={\bf K}_0^{-1} \cdot {\bf K}_i,
\end{equation}

\noindent where ${\bf K}_0$ is the global pose of the keyframe, and ${\bf K}_i$ is the global pose of the $i$-th RGB-D frame from the camera. We use the obtained transformation ${\bf T}$ to compute position of points $P=\{{\bf{p}}_1,...,{\bf{p}}_N\}$ in the keyframe coordinate system ${\bf K}_0$

\begin{equation}
    {\bf p}^0={\bf T} \cdot {\bf p}_n,
\end{equation}

\noindent where ${\bf p}^0$ is the position of the $n$-th point in the keyframe coordinate system ${K}_0$ and ${\bf p}_n$ is the position of the point in the current camera frame.

Later, an inverse model of the camera is used to compute the position of the considered point on the 2D container related to the keyframe. Then, points are grouped according to the $u$ and $v$ coordinates. Each group is used to update the ellipsoid related to cell $c_{u,v}$ in the 2D keyframe container. We use NDT update method~\cite{Saarinen2013} to update position of the ellipsoid ${\bf x}_{u,v}$ and the covariance matrix ${\bf \Sigma}_{u,v}$. The $N$ new points added to the cell and current mean value ${\bf x}_t$ computed from total $M$ points are used to update the mean position of the ellipsoid ${\bf x}_{t+1}$:

\begin{equation}
    {\bf x}_{t+1} = \frac{{\bf x}_t M + \sum_{n=1}^{N} {\bf p}_n}{M+N}.
\end{equation}

The update procedure is repeated whenever a new point cloud from the sensor is provided. The iterative procedure requires the computation of the sum $\pmb{s}_M=\sum_{m=1}^M \pmb{p}_m$ and the total number of points M used to update the considered cell (ellipsoid). We update the covariance matrix using:

\begin{multline}
 {\bf \Sigma}_{t+1} = {\bf \Sigma}_{t+1} + \sum_{n=1}^{N} \left({\bf p}_n-\frac{{\bf s}_N}{N}\right)\cdot \left({\bf p}_n-\frac{{\bf s}_N}{N}\right)' +\\
 \frac{M}{N(M+N)}\left(\frac{N}{M}{\bf s}_M - {\bf s}_N\right)\left(\frac{N}{M}{\bf s}_M - {\bf s}_N\right)',
 \label{eq:updateCov}
\end{multline}

\noindent where ${\bf s}_N=\sum_{n=1}^N {\bf p}_n$ and N is the size of point cloud used to update the voxel. To update the color of the voxel, we use:

\begin{eqnarray}
 {\bf c}_{t+1} & = & \frac{M \cdot {\bf c}_{t}+{\bf s}_{c_N}}{M+N},
 \label{eq:color_update}
\end{eqnarray}

\noindent where ${\bf s}_{c_N}=\sum_{n=1}^N {\bf c}_n$ is the sum of colors for points P used to update voxel, $\pmb{c}_t$ is the current mean color of the voxel.

\begin{figure}[t]
\centering
\includegraphics[width=0.99\columnwidth]{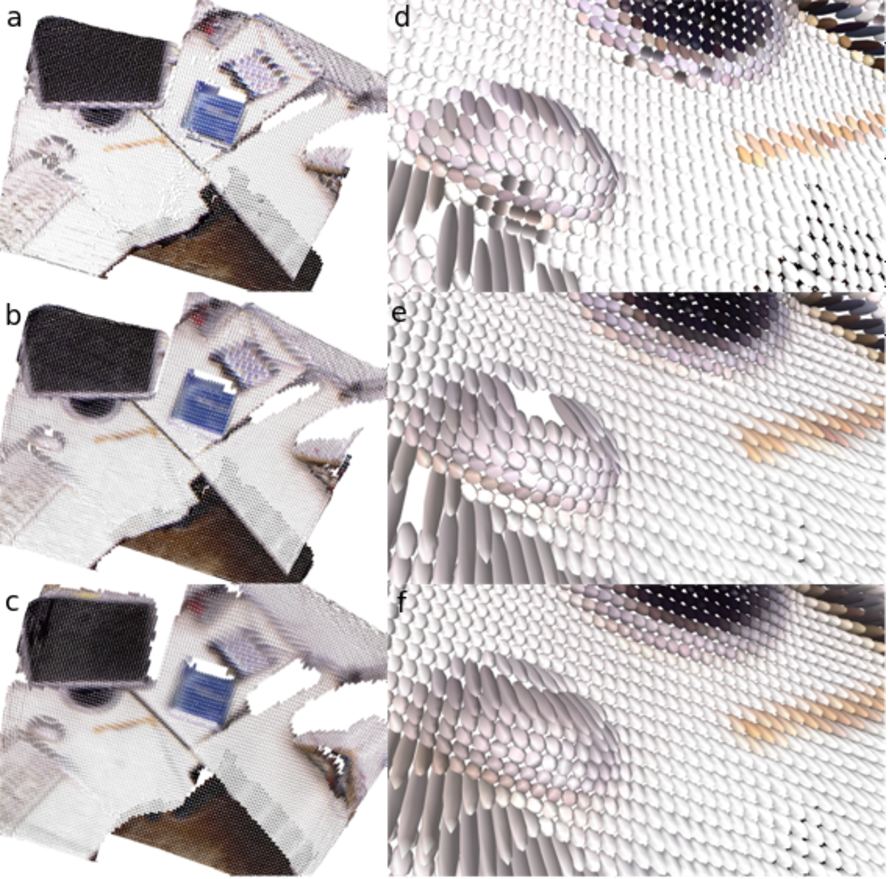}
\caption{Example local map generated using, consecutively, 1, 2 and 10 frames, obtained for the {\tt\scriptsize freiburg1\_desk} sequence from the TUM dataset~\cite{Sturm2012}: 2D container with ellipsoids (a,b,c) and the enlarged region of the local map (d,e,f)}
\label{localMap_sequence}
\end{figure}

After each cell update, the total number of points used to update single ellipsoid is increased $M=M+N$. In Fig.~\ref{localMap_sequence} we show the set of ellipsoids stored in the 2D container after 1, 2, and 10 update iterations. In the first iteration, the ellipsoids represent the real shape of the objects. The larger size of the ellipsoids naturally represents the larger uncertainty of the measurements. In the following iterations the shape described by the ellipsoids becomes more smooth (compare enlarged region in Fig.~\ref{localMap_sequence} after 1 and 10 iterations). 

\subsection{Ellipsoids filtering}

To remove incorrect ellipsoids that appear on the edges of the objects we use the uncertainty model of the RGB-D sensor. Because the ellipsoids computed using NDT updating procedure are closely related to the uncertainty model of the RGB-D sensor~\cite{Belter2018mva} we compare directly the model with the obtained ellipsoids. The filtering procedure is based on rejecting ellipsoids that are too long in 3D space according to the uncertainty characteristic of the RGB-D sensor. Ellipsoids with center points close to the camera origin should have a much smaller magnitude than those further away according to the sensor model. For results obtained from various datasets, we use the same generic uncertainty model of the Kinect sensor~\cite{Funek2018}.



\subsection{Merging local maps}
\label{merging_sub}

To compute the global map of the environment we have to compute the position of each ellipsoid from the local map in the global coordinate frame. To this end, we iterate over each local map $k_i$ and compute the global pose of each ellipsoid. With the given set of ellipsoids, we can easily obtain the global point cloud and global Octomap by sampling from the set of ellipsoids~\cite{Belter2018}. However, in contrast to our previous approach where local maps have the same resolution in the 3D space, we store local maps with different 3D accuracy. Thus, ellipsoids from different local maps which describe the same region of the 3D space have various size. We propose a procedure that merges ellipsoids from various viewpoints, to merge the information from the local maps.

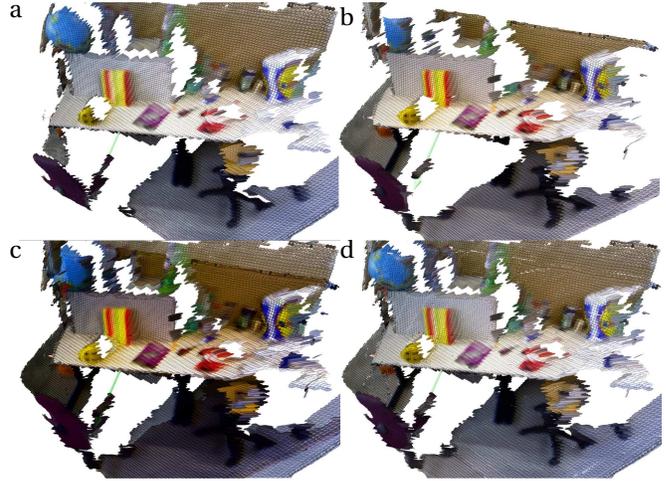
\begin{figure}[t]
\centering
\includegraphics[width=0.99\columnwidth]{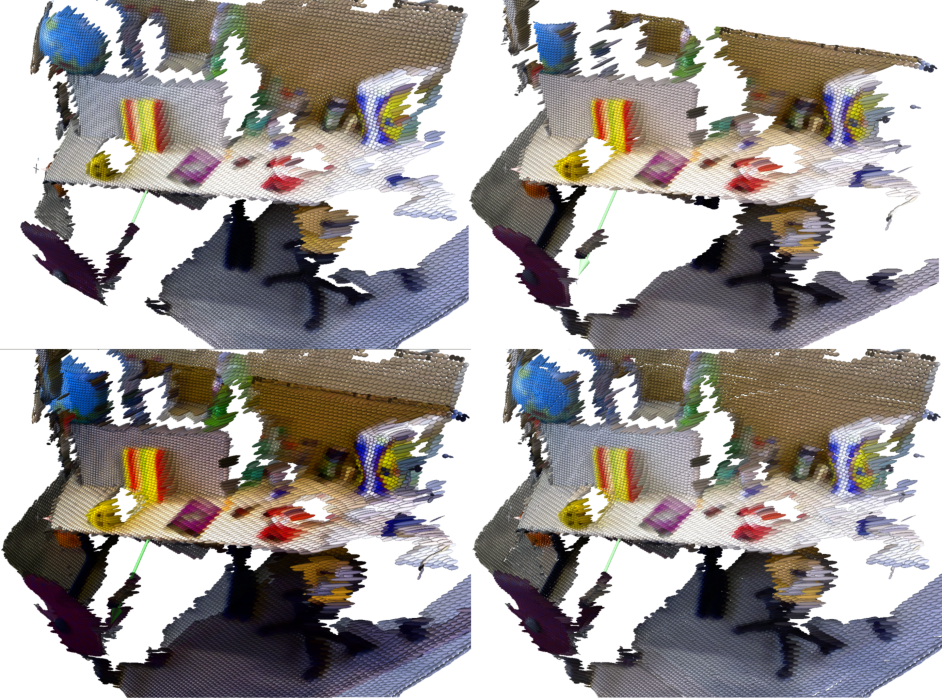}
\put(-247,175){a} \put(-122,172){b}
\put(-247,84){c} \put(-122,84){d}
\caption{Merging two local maps (a,b): maps in the common coordinate frame (c) and the obtained set of ellipsoids (d).}
\label{mapMerging}
\end{figure}

Firstly, we merge ellipsoids from neighboring local maps. The example of merging two neighboring local maps is presented in Fig.~\ref{mapMerging}. The local maps which are connected with the edges in the covisibility graph cover the same region in the 3D space. In Fig.~\ref{mapMerging}c we show how the local maps overlap. We merge ellipsoids that are close to each other in the 3D space. However, clustering in the 3D space is computationally expensive and time-consuming. Instead, we divide the local maps into the groups of neighboring maps. Then, we re-project all ellipsoids from neighboring maps to the map in the center of the group using inverse pinhole camera model~\cite{Belter2018mva}:


\begin{align}
    \begin{bmatrix} u\\
    v\\
    d\\
    \end{bmatrix}&= 
         \begin{bmatrix} \frac{x \cdot f_x}{z}+x_c\\
        \frac{y \cdot f_y}{z}+y_c\\
        z\\
    \end{bmatrix},
\end{align}

where $[u, v, d]^T$ is the position of the ellipsoid on the new image plane, $[x, y, z]^T$ is the position of the ellipsoid in the camera frame, $x_c$, $y_c$ define the position of the optical axis on image plane, $f_x$ and $f_y$ are focal lengths of the camera. Then, ellipsoids which are located in the same 2D cell $c_{u,v}$ are clustered using mean shift clustering. We do not cluster if the distance between ellipsoids is larger than the threshold (we set the threshold value to 0.25~m in all experiments presented in the paper). For all ellipsoids in the same cluster, we compute the mean position, covariance matrix, and the color. The results of the clustering are presented in Fig.~\ref{mapMerging}d. By applying the proposed procedure we reduce significantly the number of ellipsoids and smooth the global map.

\begin{figure}[t]
\centering
\includegraphics[width=0.93\columnwidth]{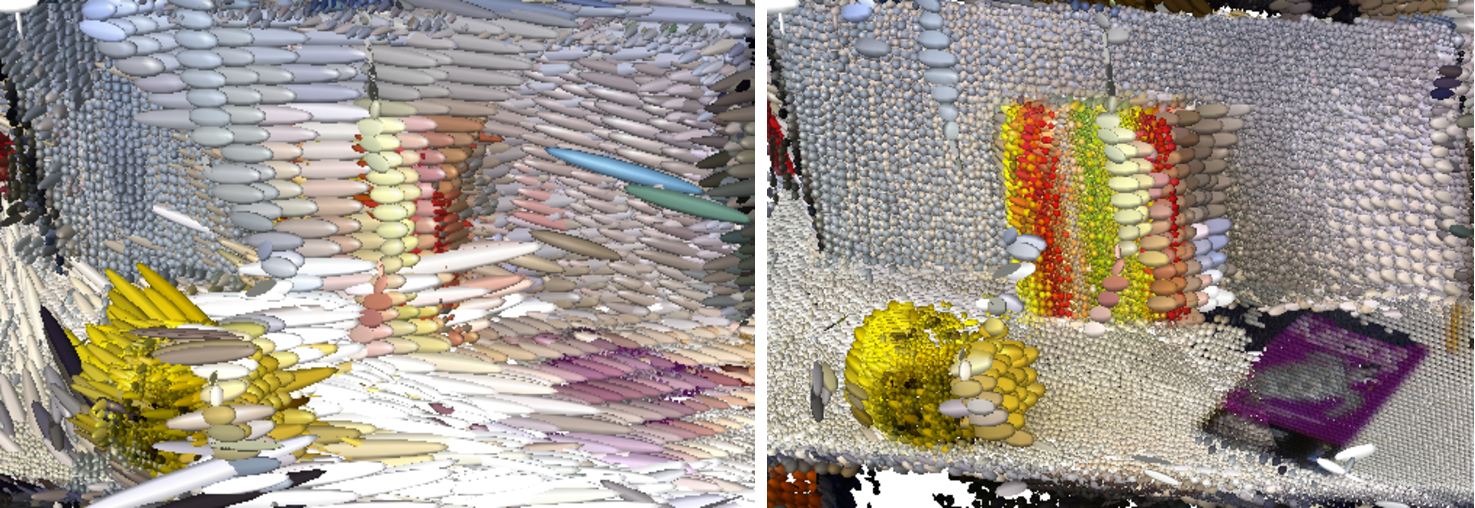}
\put(-225,70){\color{white}{a}} \put(-107,70){b}
\caption{Occluding ellipsoids removal procedure which allows keeping ellipsoids which represent objects with the highest precision: before (a) and after filtering (b)}
\label{removeOccluding}
\end{figure}

Secondly, we have to deal with the various resolution of the local maps in the 3D space. The same object can be observed from various viewpoints and included in multiple local maps. Each local map produces ellipsoids with various sizes in the 3D space. If the object is observed with various precision, our goal is to keep the most precise model. To this end, we remove ellipsoids which occlude other ellipsoids that have smaller uncertainty. We use the procedure which is similar to the procedure used to merge ellipsoids. Again, we create groups of local maps which are neighbors in the covisibility graphs but also we add local maps in which Euclidean distance and the angle between camera axes are below the threshold. We check if each ellipsoid from the local map is occluded by other ellipsoids for the given viewpoint. We can perform this operation efficiently because we apply the computationally efficient forward and inverse camera model and we store information about 2D coordinates of each ellipsoid on the image. For the map with 100 local maps, the filtering procedure takes less than 5~s. Example results of the occluding removal procedure are presented in Fig.~\ref{removeOccluding}.


\section{Results}

\begin{figure}[t]
\centering
\includegraphics[width=0.89\columnwidth]{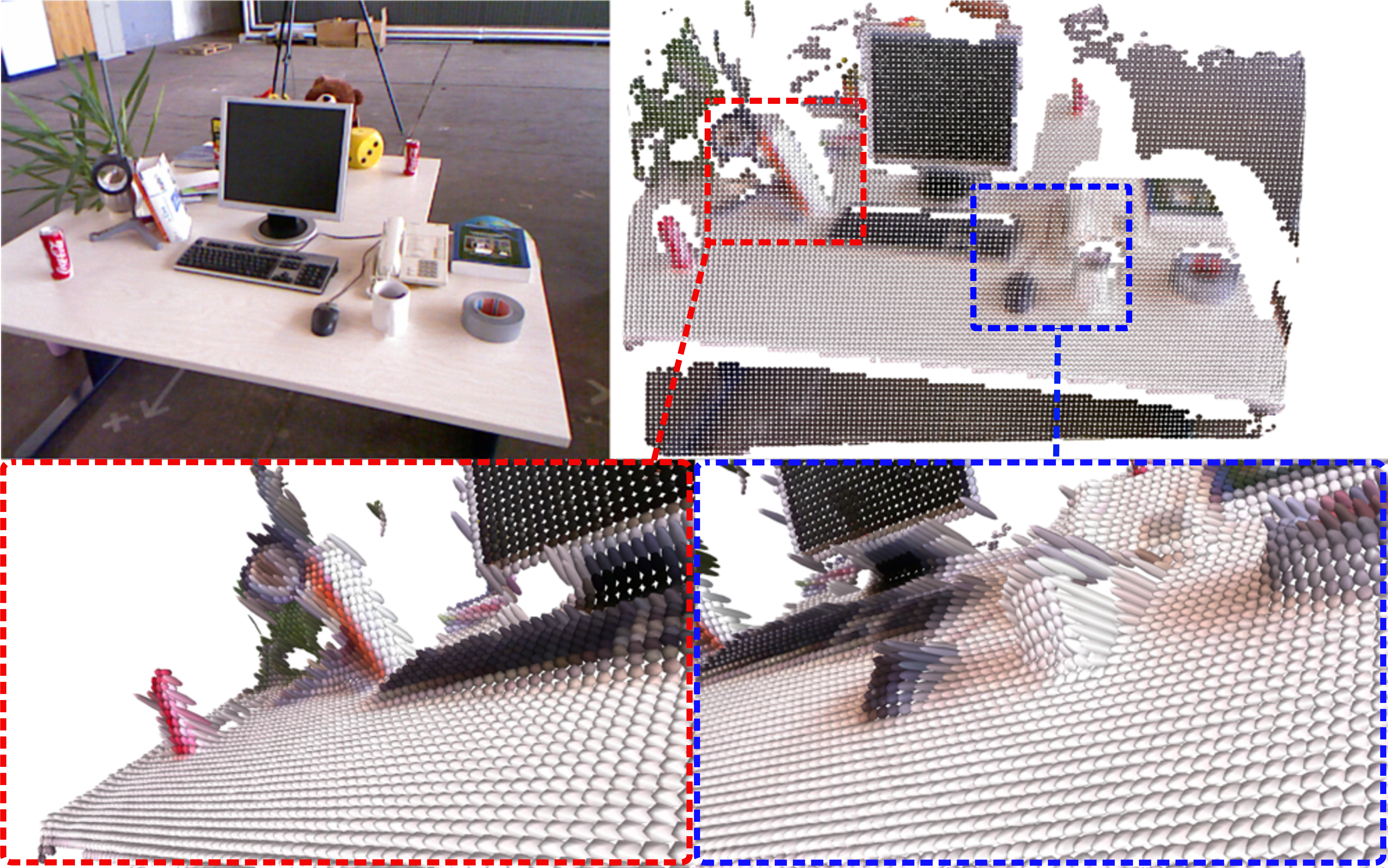}
\caption{Local map obtained for the {\tt\scriptsize freiburg2\_desk} sequence from the TUM dataset~\cite{Sturm2012}. The enlarged regions of the map are in the red and blue frames.}
\label{localMap_with_details}
\end{figure}

We performed experiments on various datasets to show the properties of the proposed method\footnote{short video from experiments is available at \url{https://youtu.be/uNEZbMYmRq4}}. In the first experiment, we present the example keyframe (local map) of the {\tt freiburg3\_desk} sequence from the TUM dataset~\cite{Sturm2012}. In Fig.~\ref{localMap_with_details} we show the set of ellipsoids stored in the 2D container. We also enlarge the selected regions of the local map to show the properties of the map. The ellipsoids which are closer to the camera are much smaller than the distant ones. On the other hand, the ellipsoids located on the edges of the objects (a cup, coke, and mouse in Fig.~\ref{localMap_with_details}) are elongated due to measurements (depth values) along the $z$ axis of the camera. Fig.~\ref{localMap_with_details} also show that the local map contains information about the object from a single viewpoint. The orientation and position of the camera do not change significantly which allows us to integrate 3D measurements on the 2D plane.



\begin{figure}[t]
\centering
\includegraphics[width=0.89\columnwidth]{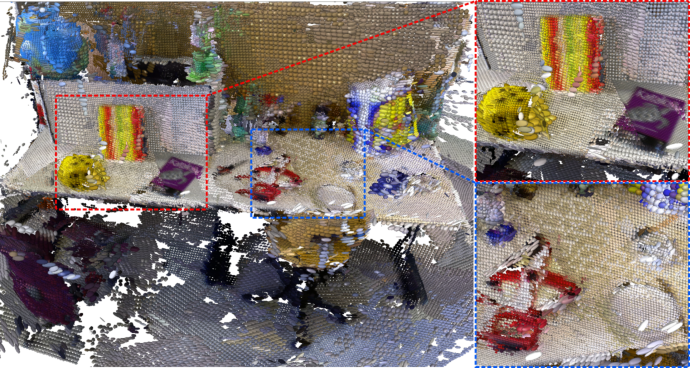}
\caption{Global map obtained for the {\tt\scriptsize fr3\_long} sequence from the TUM dataset~\cite{Sturm2012}. The enlarged regions of the map are in the red and blue frames.}
\label{globalMap}
\end{figure}

In the second experiment, we show the global map obtained for the {\tt fr3\_long} sequence from the TUM dataset~\cite{Sturm2012}. The results are presented in Fig.~\ref{globalMap}. We show the global set of ellipsoids. We applied the proposed methods of merging local maps and removing occluding ellipsoids. Finally, the global map contains 358258 ellipsoids which is less than the number of pixels in a single RGB image. In Fig.~\ref{globalMap} we also show the enlarged regions of the global map. Even though the global map is built using local maps with various 3D precision we can integrate all measurements and restore detailed information about objects.

\begin{table}[t]{
\caption{Parameters of the maps obtained on the {\tt fr3\_long\_office}, {\tt freiburg1\_room}, and {\tt\scriptsize fr2\_desk} sequences from the TUM dataset. We compare the total number of ellipsoids in the local maps $\Sigma_e$, ellipsoids after clustering and filtering $\Sigma_e'$, and time required to generate global map $t_g$}
\label{resultsTab}
\vspace*{-10pt}
\begin{center}
{\footnotesize
\begin{tabular}{l|c|c|c}
\thead{sequence}                     & \thead{{\tt\scriptsize fr3\_long}} &  \thead{{\tt\scriptsize fr1\_room}} & \thead{{\tt\scriptsize fr2\_desk}}\\
\hline
submaps no.             & 161 & 88 & 99 \\
$\Sigma_e$              & 892232 & 405137 & 390435 \\
$\Sigma_e'$          & 358258 & 189434 & 90719 \\
$t_g$ [s]             & 14.2 & 4.8 & 3.9 \\
\end{tabular}
}
\end{center}
}
\end{table}

We summarize the properties of the global maps obtained on the three sequences from the TUM dataset in Tab.~\ref{resultsTab} obtained on the computer with i5-8250U CPU. The global map presented in Fig.~\ref{globalMap} contains initially 892232 ellipsoids. After clustering and filtering, this number is reduced more than two times. A similar reduction rate is obtained for other sequences. The maximal computation time for the global map is 14.2~s which is obtained for the graph with 161 local maps. This time depends also on the total number of the ellipsoids stored in the local maps. For the smaller maps ({\tt fr2\_desk}) the global map is produced in less than 4~s.

\begin{table}[]
\caption{Comparison of OctoMap, NDT-OM and VD where $d$ - voxel size [m], $t_u$ - update time [$ms$], $\Sigma_v$ - no. of updated voxels/3D ellipsoids, 1 - OctoMap, 2 - NDT-OM, 3 - VD.}
\label{resultTab2}
\vspace*{-10pt}
\begin{center}
\setlength\tabcolsep{2.5pt}
\begin{tabular}{l|ccc|ccc|ccc}
\multirow{2}{*}{\thead{seq.}} & \multicolumn{3}{c|}{\thead{{\tt\scriptsize fr3\_long}}} & \multicolumn{3}{c|}{\thead{{\tt\scriptsize fr1\_room}}} & \multicolumn{3}{c}{\thead{{\tt\scriptsize fr2\_desk}}} \\
 & 1 & 2 & 3 & 1 & 2 & 3 & 1 & 2 & 3 \\ \hline
$d [m]$ & 0.1 & 0.1 & - & 0.1 & 0.1 & - & 0.1 & 0.1 & - \\
$t_u [ms]$ & 381 & 2275 & 1706 & 203 & 2057 & 1823 & 263 & 2182 & 1693 \\
$\Sigma_v$ & 73404 & 2255 & 358258 & 46967 & 637 & 189434 & 74527 & 1998 & 90719
\end{tabular}
\end{center}
\end{table}

\begin{figure}[t]
\centering
\includegraphics[width=0.99\columnwidth]{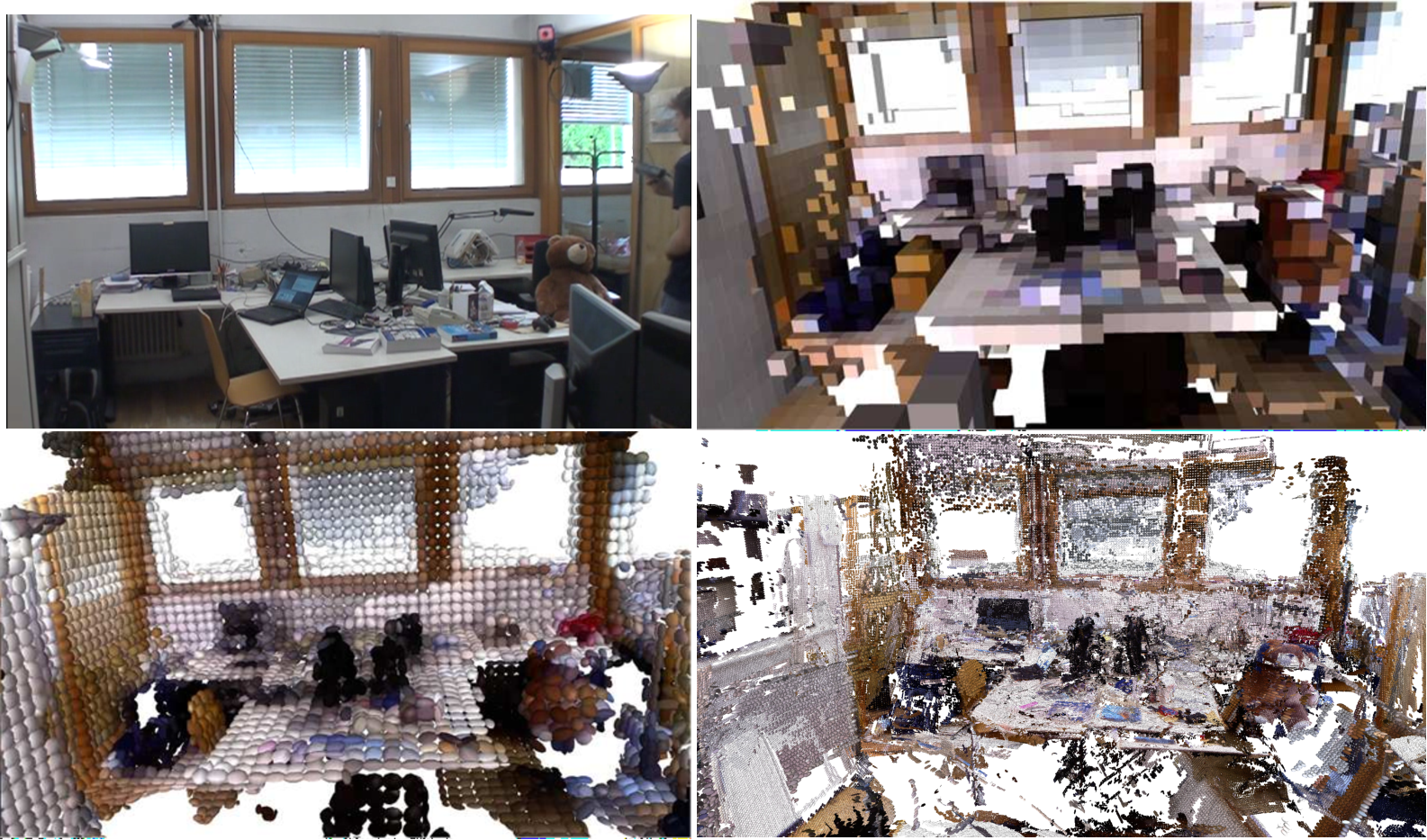}
\put(-240,134){\color{white}{a}} \put(-123,132){b}
\put(-242,62){c} \put(-123,61){d}
\caption{Global maps obtained for the {\tt\scriptsize freiburg1\_room} sequence from the TUM dataset~\cite{Sturm2012}: example RGB image (a), Octomap (b), NDT-OM (c) and our approach (d)}
\label{globalMap_comparison}
\end{figure}
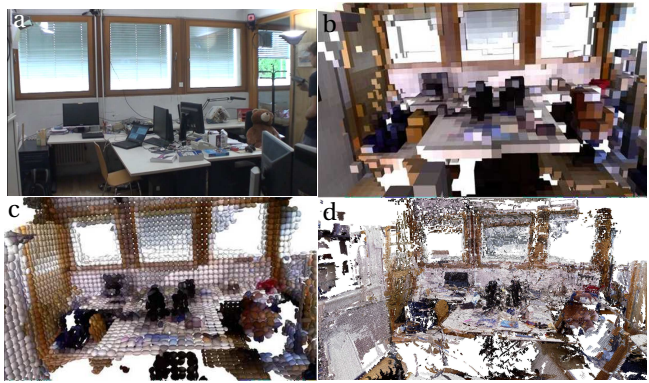

In the next experiment, we compare visually three different mapping methods: OctoMap~\cite{Hornung2013}, NDT-OM, and view-dependent approach. The used sequence here is {\tt freiburg1\_room} from the TUM dataset~\cite{Sturm2012}. In Fig.~\ref{globalMap_comparison}a, we show the experimental set. The obtained environment models are presented in Fig.~\ref{globalMap_comparison}b-d for the Octomap, NDT-OM and our approach, respectively. The Octomap and NDT-OM are generated for the voxel size equal to 0.1~m. Further reduction of the voxel size significantly increases the number of updated voxels and computation time. However, the shape of the objects is better represented by NDT-OM than Octomap. Unfortunately, this also result in higher computational cost (Tab.~\ref{resultTab2}). Our approach, presented in~\autoref{globalMap_comparison}d, provides a 3D map with varying resolution. The details of the objects included in the map are much better visible. In Tab.~\ref{resultTab2} we compare the parameters of the global maps obtained using Octomap, NDT-OM and the proposed approach. 

\begin{table}[]
\caption{Comparison of OctoMap and proposed VD approach in the reconstruction task.}
\label{resultReconsruct}
\vspace*{-10pt}
\begin{center}
\setlength\tabcolsep{3.5pt}
\begin{tabular}{l|ccc|ccc|ccc}
\multirow{2}{*}{\thead{}} & \multicolumn{3}{c|}{\thead{{\tt\scriptsize Octomap}}} & \multicolumn{3}{c|}{\thead{{\tt\scriptsize VD}}} \\
 & 0.1~m & 0.05~m & 0.02~m & 3$\times$3~px & 9$\times$9~px & 15$\times$15~px \\ \hline
$RMSE [mm]$ & 80.0 & 72.8 & 48.1 & 9.7 & 9.9 & 12.5  \\
$\Sigma_v$ & 20967 & 115441 & 1015243 & 615206 & 63605 & 27081  \\
\end{tabular}
\end{center}
\end{table}

The comparison between OctoMap~\cite{Hornung2013} and the proposed method in the reconstruction task is presented in Tab.~\ref{resultReconsruct}. The model of the environment is obtained on the ICL-NUIM {\tt living\_room} dataset. We compare the obtained model with the reference mesh model. To this end, we create a point cloud from the OctoMap and 3D ellipsoids by taking the centers of these geometric structures. As seen in Tab~\ref{resultReconsruct}, when the size of the voxel is decreased, we improve the model of the environment but the number of voxels significantly increases (to 1015243 when the voxel size is 0.02~m). The RMSE for the OctoMap with 0.02~m voxel size is relatively high (48.1 mm). Further decreasing the voxel size increase significantly the memory consumption. The reconstruction error is much smaller when the proposed view-dependent model is used. Even when the cell size is set to 15$\times$15~px the RMSE is 12.5~mm and the number of ellipsoids is 27081 only.

\section{Conclusions and Future Work}
\noindent

In this article, we present the mapping method which is based on the set of view-dependent local maps connected by the edges in the pose graph. The local maps allow integrating measurements from the RGB-D frames which are close to the pose of the keyframe. The obtained local model represents precisely objects which are close to the camera pose. The distant objects are represented in the local map with lower accuracy. This approach is closely related to the uncertainty model of the 3D sensors. With the proposed approach we do not lose information about objects which are smaller than the given threshold like the voxel size in the Octomap and NDT-OM mapping methods. Moreover, we can correct the global map when the loop closure is detected~\cite{Belter2018}. 

In this research, we also compare the proposed approach with the Octomap and NDT-OM. We generate a global map to show the properties and capabilities of the method to represent the environment with various resolutions. However, we are focused on the application of local maps. The local representation of the environment is sufficient to plan the motion of the arm of the mobile manipulation robot. The local map can be also used to detect objects and localize the robot by matching the current data from the sensor to the set of local maps (keyframes) like it is implemented in the ORB-SLAM2~\cite{Mur-Artal2017}. In the future, we are going to integrate ORB-SLAM2 with our keyframe-based dense mapping method and share directly data structures between the localization system and dense mapping method. We are also going to work on the simultaneous integration of the local maps in the global map and camera pose estimation.



\end{document}